\documentclass[10pt,twocolumn,letterpaper]{article}

\usepackage[pagenumbers]{cvpr} % To force page numbers, e.g. for an arXiv version

\usepackage{graphicx}
\usepackage{amsmath}
\usepackage{amssymb}
\usepackage{booktabs}
\usepackage{enumitem}
\usepackage{booktabs}
\usepackage{multirow}
\usepackage{pifont}
\usepackage{xcolor}
\usepackage{subcaption}
\usepackage{wasysym}
\usepackage{comment}
\usepackage{pifont}
\usepackage{amssymb}
\usepackage{multirow}
\usepackage{makecell}
\usepackage{caption}
\usepackage{soul}

\usepackage[pagebackref,breaklinks,colorlinks]{hyperref}

\usepackage[capitalize]{cleveref}

\crefformat{section}{\S#2#1#3} % see manual of cleveref, section 8.2.1
\crefformat{subsection}{\S#2#1#3}
\crefformat{subsubsection}{\S#2#1#3}

\crefname{section}{Sec.}{Secs.}
\Crefname{section}{Section}{Sections}
\Crefname{table}{Table}{Tables}
\crefname{table}{Tab.}{Tabs.}

\def\cvprPaperID{*****}
\def\confName{CVPR}
\def\confYear{2022}

\newcommand{\model}{OpenELM\@\xspace}

\definecolor{deemph}{gray}{0.6}
\newcommand{\gc}[1]{\textcolor{deemph}{#1}}

\definecolor{yes}{RGB}{51,160,44}
\definecolor{no}{RGB}{228,26,28}

\newcommand{\cmark}{\textcolor{yes}{\ding{51}}}%
\newcommand{\xmark}{\textcolor{no}{\ding{55}}}%

\begin{document}

%%%%%%%%% TITLE - PLEASE UPDATE
\title{\model: An Efficient Language Model Family with Open Training and Inference Framework}

\author{Sachin Mehta \qquad Mohammad Hossein Sekhavat \qquad Qingqing Cao \qquad Maxwell Horton \\ Yanzi Jin \qquad Chenfan Sun \qquad Iman Mirzadeh \qquad Mahyar Najibi \qquad Dmitry Belenko \\  Peter Zatloukal \quad Mohammad Rastegari \\
Apple\\
}

\twocolumn[{%
\renewcommand\twocolumn[1][]{#1}%
\maketitle
\begin{center}
    \centering
    \captionsetup{type=table}
    \resizebox{1.5\columnwidth}{!}{
    \begin{tabular}{lcccccc}
        \toprule[1.5pt]
        \multicolumn{1}{c}{\multirow{2}{*}{\textbf{Model}}} & \multicolumn{1}{c}{\multirow{2}{*}{\textbf{Public dataset}}} & \multicolumn{2}{c}{\textbf{Open}}& \multicolumn{1}{c}{\multirow{2}{*}{\textbf{Model size}}} & \multicolumn{1}{c}{\multirow{2}{*}{\textbf{Pre-training tokens}}} & \multicolumn{1}{c}{\multirow{2}{*}{\textbf{Average acc. (in \%)}}} \\
        \cmidrule[1.25pt]{3-4}
         & & \textbf{Code} & \textbf{Weights} & & & \\
        \midrule[1.25pt]
        \gc{OPT} \cite{zhang2022opt} & \xmark & \cmark & \cmark & \gc{1.3 B} & \gc{0.2 T} & \gc{41.49} \\
        \gc{PyThia} \cite{biderman2023pythia} & \cmark & \cmark & \cmark & \gc{1.4 B} & \gc{0.3 T} & \gc{41.83} \\ 
        MobiLlama \cite{thawakar2024mobillama} & \cmark & \cmark & \cmark & 1.3 B & 1.3 T  & 43.55 \\ 
        OLMo \cite{groeneveld2024olmo} & \cmark & \cmark & \cmark & 1.2 B & 3.0 T & 43.57 \\  
        \model (Ours) & \cmark & \cmark & \cmark & 1.1 B & 1.5 T  & \textbf{45.93} \\ 
        \bottomrule[1.5pt]
    \end{tabular}
    }
    \captionof{table}{\textbf{\model vs. public LLMs.} \model outperforms comparable-sized existing LLMs pretrained on publicly available datasets. Notably, \model outperforms the recent open LLM, OLMo, by 2.36\% while requiring $2\times$ fewer pre-training tokens. The average accuracy is calculated across multiple tasks listed in \cref{tab:openllm}, which are also part of the OpenLLM leaderboard \cite{open-llm-leaderboard}. Models pretrained with less data are highlighted in \gc{gray} color.}
    \label{tab:compare_sota_teaser}
\end{center}%
}]

%%%%%%%%% ABSTRACT
\begin{abstract}
The reproducibility and transparency of large language models are crucial for advancing open research, ensuring the trustworthiness of results, and enabling investigations into data and model biases, as well as potential risks. To this end, we release \model, a state-of-the-art open language model. \model uses a layer-wise scaling strategy to efficiently allocate parameters within each layer of the transformer model, leading to enhanced accuracy. 
For example, with a parameter budget of approximately one billion parameters, \model exhibits a 2.36\% improvement in accuracy compared to OLMo while requiring $2\times$ fewer pre-training tokens.

Diverging from prior practices that only provide model weights and inference code, and pre-train on private datasets, our release includes the complete framework for training and evaluation of the language model on publicly available datasets, including training logs, multiple checkpoints, and pre-training configurations. We also release code to convert models to MLX library for inference and fine-tuning on Apple devices. This comprehensive release aims to empower and strengthen the open research community, paving the way for future open research endeavors. 

Our source code along with pre-trained model weights and training recipes is available at \url{https://github.com/apple/corenet}. Additionally, \model models can be found on HuggingFace at: \url{https://huggingface.co/apple/OpenELM}.
\end{abstract}

%%%%%%%%% BODY TEXT
\section{Introduction}
\label{sec:intro}

Transformer-based \cite{vaswani2017attention} large language models (LLM) are revolutionizing the field of natural language processing \cite{brown2020language,touvron2023llama}. These models are isotropic, meaning that they have the same configuration (\textit{e.g.}, number of heads and feed-forward network dimensions) for each transformer layer. Though such isotropic models are simple, they may not allocate parameters efficiently inside the model. 

In this work, we develop and release \model, a family of pre-trained and fine-tuned models on \emph{publicly} available datasets. At the core of \model lies layer-wise scaling \cite{mehta2020delight}, enabling \emph{more efficient parameter} allocation across layers. This method utilizes smaller latent dimensions in the attention and feed-forward modules of the transformer layers closer to the input, and gradually widening the layers as they approach the output.

We release the complete framework, encompassing data preparation, training, fine-tuning, and evaluation procedures, alongside multiple pre-trained checkpoints and training logs, to facilitate open research. Importantly, \model outperforms existing open LLMs that are pre-trained using publicly available datasets (\cref{tab:compare_sota_teaser}). For example, \model with 1.1 billion parameters outperforms OLMo \cite{groeneveld2024olmo}, which has 1.2 billion parameters, by 2.36\% while requiring $2\times$ fewer pre-training tokens.

\section{Pre-training}
This section describes the framework, including model architecture (\cref{ssec:model_arch}), pre-training data (\cref{ssec:pretrain_data}), training hyper-parameters (\cref{ssec:training_hyper}), and evaluation (\cref{ssec:evaluation_details}).

\subsection{\model architecture}
\label{ssec:model_arch}
We adopt the decoder-only transformer-based architecture. Following state-of-the-art LLMs, we: (1) do not use learnable bias parameters in any fully-connected (\textit{a.k.a.}, linear) layers, (2) apply pre-normalization using RMSNorm \cite{zhang2019root} and also, use rotatory positional embedding (ROPE) \cite{su2024roformer} for encoding positional information, (3) use grouped query attention (GQA) \cite{ainslie2023gqa} instead of multi-head attention (MHA), (4) replace the feed forward network (FFN)  with SwiGLU FFN \cite{shazeer2020glu}, (5) use flash attention \cite{dao2022flashattention} for computing the scaled dot-product attention, and (6) use the same tokenizer as LLama \cite{touvron2023llama}.

Existing LLMs use the same configuration for each transformer layer in the model, resulting in a uniform allocation of parameters across layers. Unlike these models, each transformer layer in \model has a different configuration (\textit{e.g.}, number of heads and feed forward network dimension), resulting in variable number of parameters in each layer of the model. This lets \model to better utilize the available parameter budget for achieving higher accuracies. We implement this non-uniform allocation of parameters across layers using \emph{layer-wise scaling} (also referred as block-wise scaling in \cite{mehta2020delight}).

\paragraph{Layer-wise scaling.} A standard transformer layer is composed of multi-head attention (MHA) and feed-forward network (FFN). For non-uniform allocation of parameters in the transformer layer, we adjust the number of attention heads and the FFN multiplier in each transformer layer.

Assume that the standard transformer model with uniform parameter allocation has $N$ transformer layers and the dimensionality of the input to each layer is $d_{model}$. The MHA has $n_h$ heads and dimension of each head is $d_h =\frac{d_{model}}{n_h}$. Also, the hidden dimension for FFN is $d_{\textrm{FFN}} = m \cdot d_{model}$, where $m$ is a scalar FFN multiplier. 

We introduce parameters $\alpha$ and $\beta$ to scale the number of attention heads $n_h$ and FFN multiplier $m$ per layer respectively. For the $i$-th layer, $n_h$ and $m$ are computed as
\begin{equation}
    \begin{split}
    n^i_h & = \frac{\alpha^i \cdot d_{model} }{d_h}, \quad m^i = \beta^i \\
    \text{where } & \alpha^i = \alpha_{min} + \frac{(\alpha_{max} - \alpha_{min})\cdot i}{N - 1}, \\ \text{and } & \beta^i = \beta_{min} + \frac{(\beta_{max} - \beta_{min}) \cdot i}{N - 1}, 0 \le i < N.
    \end{split}
    \label{eq:layer_wise_scale}
\end{equation}
Here, $\alpha_{min}$ and $\alpha_{max}$ are the hyper-parameters that allow us to scale the attention heads. Similarly, $\beta_{min}$ and $\beta_{max}$ let us to vary the width of FFN layers. Therefore, varying the configuration of standard transformer layers using $\alpha$ and $\beta$ results in non-uniform allocation of parameters in the model. Note, setting $\alpha_{min} = \alpha_{max} = 1.0$ and $m_i=m$ produces the standard uniform transformer model.

\subsection{Pre-training data}
\label{ssec:pretrain_data}

For pre-training, we use public datasets. Specifically, our pre-training dataset contains RefinedWeb \cite{penedo2023refinedweb}, deduplicated PILE \cite{gao2020pile}, a subset of RedPajama \cite{together2023redpajama}, and a subset of Dolma v1.6 \cite{soldaini2024dolma}, totaling approximately 1.8 trillion tokens. These details are also summarized in \cref{tab:dataset_mix_pretrainig}.

\paragraph{On-the-fly tokenization and data filtering.} Unlike previous approaches that utilize pre-tokenized data \cite{biderman2023pythia,groeneveld2024olmo}, we filter and tokenize text data on-the-fly. This facilitates seamless experimentation with various tokenizers, thereby significantly simplifying prototyping and research endeavors. In our experiments, we use the same tokenizer as used in LLama \cite{touvron2023llama}.

To filter out low-length sequences, we apply two filtering methods. The first method operates at the \emph{character-level}, checking if the number of characters in the sequence is below a specified threshold. The second method operates at the \emph{token-level}, where it examines whether the sequence contains fewer tokens than a specified threshold. Sequences that are shorter than either of these thresholds are skipped. In our experiments, we use 200 characters and 256 tokens as character and token-level filtering thresholds.

\begin{table}[b!]
    \centering
    \resizebox{0.75\columnwidth}{!}{
    \begin{tabular}{llr}
        \toprule[1.5pt]
      \textbf{Source}  & \textbf{Subset} & \textbf{Tokens} \\
      \midrule[1.25pt]
       RefinedWeb & & 665 B \\
       \midrule
       \multirow{6}{*}{RedPajama} & Github & 59 B \\
        & Books & 26 B \\
        & ArXiv & 28 B \\
        & Wikipedia & 24 B \\
        & StackExchange & 20 B \\
        & C4 & 175 B \\
        \midrule
        PILE & & 207 B\\
        \midrule 
        \multirow{5}{*}{Dolma} & The Stack & 411 B \\
              & Reddit & 89 B \\
              & PeS2o & 70 B \\
             & Project Gutenberg & 6 B \\
             & Wikipedia + Wikibooks & 4.3 B \\
        \bottomrule[1.5pt]
    \end{tabular}
    }
    \caption{Dataset used for pre-training \model.}
    \label{tab:dataset_mix_pretrainig}
\end{table}

\begin{table*}[t!]
\centering
\begin{subtable}[b]{0.5\columnwidth}
    \centering
    \resizebox{0.8\columnwidth}{!}{
    \begin{tabular}{lc}
    \toprule[1.5pt]
    \textbf{Task} & \textbf{Metric} \\
    \midrule[1.25pt]
    ARC-c & Normalized accuracy \\
     ARC-e & Normalized accuracy  \\
     BoolQ & Accuracy  \\
     HellaSwag & Normalized accuracy \\
     PIQA & Normalized accuracy \\
     SciQ & Accuracy \\
     WinoGrande & Accuracy \\
    \bottomrule[1.5pt]
    \end{tabular}
}
\caption{Standard zero-shot metrics}
\label{tab:std_metrics}
\end{subtable}
\hfill
\begin{subtable}[b]{0.75\columnwidth}
    \centering
    \resizebox{\columnwidth}{!}{
    \begin{tabular}{lcc}
     \toprule[1.5pt]
    \multirow{2}{*}{\textbf{Task}} & \multirow{2}{*}{\textbf{Metric}} & \textbf{Num. few} \\
    &  & \textbf{shot examples} \\
    \midrule[1.25pt]
    ARC-c & Normalized accuracy & 25 \\
    HellaSwag & Normalized accuracy & 10 \\
    MMLU & Accuracy & 5 \\
    TruthfulQA-mc2 & Accuracy & 0 \\
    WinoGrande & Accuracy & 5 \\
    \bottomrule[1.5pt]
    \end{tabular}
}
\caption{OpenLLM leaderboard}
\label{tab:openllm}
\end{subtable}
\hfill
\begin{subtable}[b]{0.6\columnwidth}
    \centering
    \resizebox{\columnwidth}{!}{
    \begin{tabular}{lcc}
     \toprule[1.5pt]
    \multirow{2}{*}{\textbf{Task}} & \multirow{2}{*}{\textbf{Metric}} & \textbf{Num. few} \\
     &  & \textbf{shot examples} \\
    \midrule[1.25pt]
     ARC-c & Normalized accuracy & 25 \\
     CrowsPairs-En & PCT stereotype & 25  \\
     HellaSwag & Normalized accuracy & 10 \\
     WinoGrande & Accuracy & 5 \\
     MMLU & Accuracy & 5 \\
     PIQA & Normalized accuracy & 0 \\
     RACE & Accuracy & 0 \\
     \bottomrule[1.5pt]
     \end{tabular}
}
\caption{LLM360}
\label{tab:llm360}
\end{subtable}
\caption{Tasks and metrics used for evaluating \model.}
\label{tab:eval_metrics}
\end{table*}
\begin{figure*}[t!]
    \centering
    \begin{subfigure}[b]{0.45\columnwidth}
        \centering
        \includegraphics[width=0.95\columnwidth]{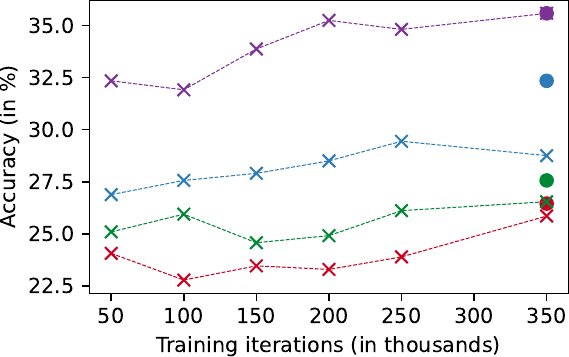}
        \caption{ARC-c}
        \label{fig:res_arc_c}
    \end{subfigure}
    \hfill
    \begin{subfigure}[b]{0.45\columnwidth}
        \centering
        \includegraphics[width=0.95\columnwidth]{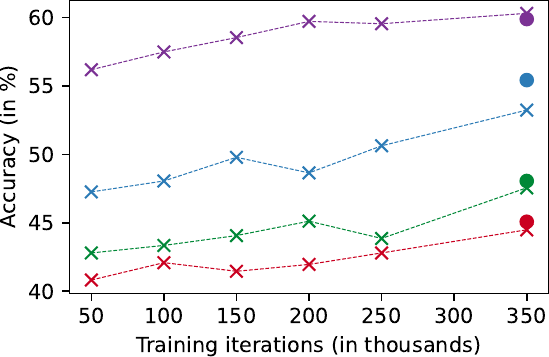}
        \caption{ARC-e}
        \label{fig:res_arc_e}
    \end{subfigure}
    \hfill
    \begin{subfigure}[b]{0.45\columnwidth}
        \centering
        \includegraphics[width=0.95\columnwidth]{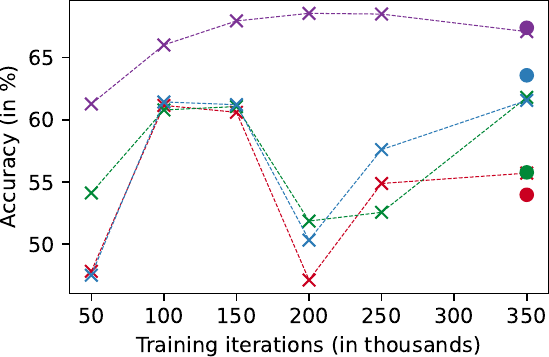}
        \caption{BoolQ}
        \label{fig:res_boolq}
    \end{subfigure}
    \hfill
    \begin{subfigure}[b]{0.45\columnwidth}
        \centering
        \includegraphics[width=0.95\columnwidth]{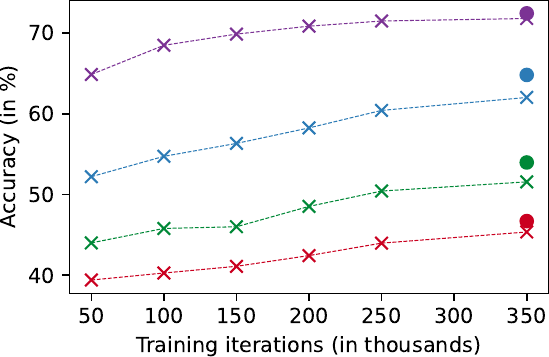}
        \caption{HellaSwag}
        \label{fig:res_hellaswag}
    \end{subfigure}
    \vfill
    \begin{subfigure}[b]{0.45\columnwidth}
        \centering
        \includegraphics[width=0.95\columnwidth]{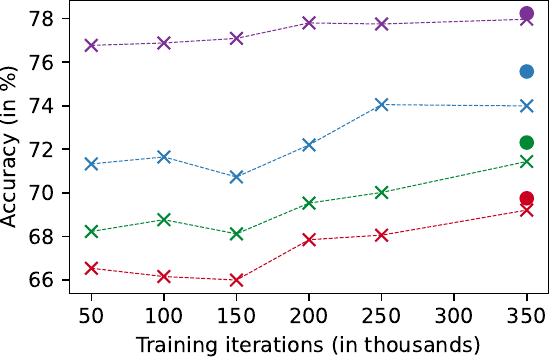}
        \caption{PIQA}
        \label{fig:res_piqa}
    \end{subfigure}
    \hfill
    \begin{subfigure}[b]{0.45\columnwidth}
        \centering
        \includegraphics[width=0.95\columnwidth]{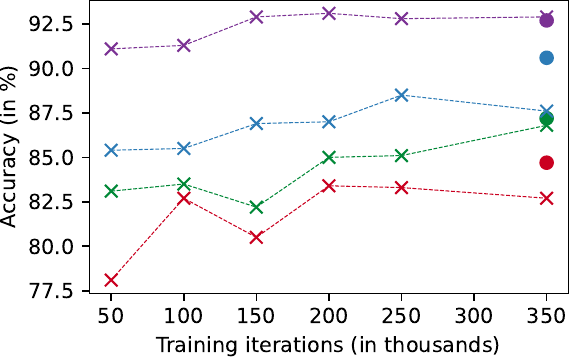}
        \caption{SciQ}
        \label{fig:res_sciq}
    \end{subfigure}
    \hfill
    \begin{subfigure}[b]{0.45\columnwidth}
        \centering
        \includegraphics[width=0.95\columnwidth]{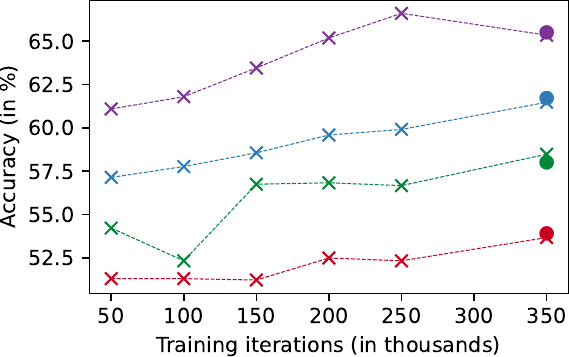}
        \caption{WinoGrande}
        \label{fig:res_winogrande}
    \end{subfigure}
    \hfill
    \begin{subfigure}[b]{0.45\columnwidth}
        \centering
        \includegraphics[width=0.5\columnwidth]{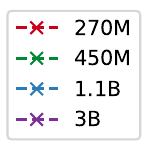}
        \caption*{\model sizes}
    \end{subfigure}
    \caption{\textbf{\model's performance across training iterations on standard zero-shot tasks}. In the majority of tasks, the performance of \model shows improvement with increasing training duration. Furthermore, the model checkpoint obtained by averaging the last five checkpoints, collected at intervals of 5k iterations, demonstrates comparable or slightly better performance (indicated by $\CIRCLE$ markers) as compared to the last checkpoint obtained after 350k iterations.}
    \label{fig:train_prog}
\end{figure*}

\subsection{Training details}
\label{ssec:training_hyper}

We train \model variants for 350k iterations (or training steps) using CoreNet (formerly CVNets \cite{mehta2022cvnets}). We use AdamW \cite{loshchilov2017decoupled} as an optimizer. We use a cosine learning rate schedule \cite{loshchilov2016sgdr}, with warm up of 5k iterations, and decay the final learning rate down to 10\% of maximum learning rate. We use a weight decay of 0.1 and gradient clipping of 1.0. We train four variants of \model (270M, 450M, 1.1B, and 3B), and for some, we use FSDP \cite{zhao2023pytorch} and activation checkpointing \cite{chen2016training}. Please refer to \cref{sec:appendix_pretrain_hyper} for additional pre-training details.

\subsection{Evaluation details}
\label{ssec:evaluation_details}
Following previous works, we evaluate the performance across different tasks using LM Evaluation Harness \cite{eval-harness}\footnote{We use commit dc90fec of \url{https://github.com/EleutherAI/lm-evaluation-harness}}:
\begin{itemize}[leftmargin=*]
    \item \textbf{Standard zero-shot tasks.} We consider 7 standard common-sense reasoning tasks: ARC easy and challenge \cite{clark2018think}, BoolQ \cite{clark2019boolq}, HellaSwag \cite{zellers2019hellaswag}, PIQA \cite{bisk2020piqa}, SciQ \cite{welbl2017crowdsourcing}, and WinoGrande \cite{sakaguchi2021winogrande}. 
    \item \textbf{OpenLLM leaderboard tasks.} We use 5 tasks from OpenLLM leaderboard \cite{open-llm-leaderboard}: ARC challenge, HellaSwag, MMLU \cite{hendrycks2020measuring}, TruthfulQA \cite{lin2021truthfulqa}, and WinoGrande.
    \item \textbf{LLM360 leaderboard tasks.} We use 7 tasks from LLM360 leaderboard \cite{liu2023llm360} for evaluation: ARC challenge, CrowS-Pairs (English version) \cite{nangia2020crows}, HellaSwag, WinoGrande, MMLU, PIQA, and RACE \cite{lai2017race}.
\end{itemize}
These evaluation frameworks, built on top of LM Evaluation Harness, allows us to comprehensively evaluate \model in terms of reasoning (\textit{e.g.}, ARC-c, HellaSwag, and PIQA), knowledge understanding (\textit{e.g.}, MMLU and RACE), and misinformation \& bias (\textit{e.g.}, TruthfulQA and CrowS-Pairs). While there may be some overlap in tasks among these frameworks, they primarily differ in the few-shot settings, as outlined in \cref{tab:eval_metrics}.

\begin{table*}[t!]
    \centering
\begin{subtable}[c]{2\columnwidth}
    \centering
    \resizebox{\columnwidth}{!}{
\begin{tabular}{l|cc|ccccccc|cc}
\toprule[1.5pt]
\textbf{Model} & \textbf{Model size} & \textbf{Pretraining tokens}   & \textbf{ARC-c} & \textbf{ARC-e} & \textbf{BoolQ} & \textbf{HellaSwag} & \textbf{PIQA}  & \textbf{SciQ}  & \textbf{WinoGrande} & \textbf{Average} & \textbf{Average w/o SciQ}\\ 
\midrule[1.25pt]
\model (Ours) & 0.27 B & 1.5 T & 26.45 & 45.08 & 53.98 & 46.71     & 69.75 & 84.70 & 53.91      & 54.37  &    49.31        \\
\hline
MobiLlama \cite{thawakar2024mobillama} & 0.50 B  & 1.3 T   & 26.62 & 46.04 & 55.72 & 51.06     & 71.11 & 83.60 & 53.20      & 55.34   &    50.63       \\
\model (Ours) & 0.45 B & 1.5 T & \textbf{27.56} & \textbf{48.06} & \textbf{55.78} & \textbf{53.97} & \textbf{72.31} & \textbf{87.20} & \textbf{58.01}  & \textbf{57.56}      &    \textbf{52.62}    \\
\hline
TinyLlama \cite{zhang2024tinyllama} & 1.10 B & 3.0 T & 30.12 & 55.25 & 57.83 & 59.20     & 73.29 & -     & 59.12      & -       &   55.80    \\
OpenLM \cite{openlm} & 1.00 B & 1.6 T & 31.00 & 56.00 & \textbf{65.00} & 61.00    & 74.00 & -     & 60.00      & -  &   57.83         \\
MobiLlama \cite{thawakar2024mobillama} & 0.80 B & 1.3 T & 28.84 & 49.62 & 60.03 & 52.45     & 73.18 & 85.90 & 55.96      & 58.00      &   53.35     \\
MobiLlama \cite{thawakar2024mobillama} & 1.26 B & 1.3 T & 31.91 & 56.65 & 60.34 & 62.18     & 74.81 & 89.10 & 59.27      & 62.04      &    57.53    \\
OLMo \cite{groeneveld2024olmo} & 1.18 B & 3.0 T & 31.06 & \textbf{57.28} & 61.74 & 62.92     & 75.14 & 87.00 & 59.98      & 62.16  &    58.02        \\
\model (Ours) & 1.08 B & 1.5 T   & \textbf{32.34} & 55.43 & 63.58 & \textbf{64.81}     & \textbf{75.57} & \textbf{90.60} & \textbf{61.72} & \textbf{63.44}   &  \textbf{58.91}         \\
\hline
\model (Ours) & 3.04 B & 1.5 T   & 35.58 & 59.89 & 67.40 & 72.44     & 78.24 & 92.70 & 65.51      & 67.39    &  63.18        \\
\bottomrule[1.5pt]

\end{tabular}
}
\caption{Results on zero-shot tasks with respect to the standard metrics defined in \cref{tab:std_metrics}.}
\label{tab:res_std_metrics}
\end{subtable}
\vfill
\begin{subtable}[c]{2\columnwidth}
    \centering
    \resizebox{0.95\columnwidth}{!}{
        \begin{tabular}{l|cc|ccccc|c}
            \toprule[1.5pt]
            \textbf{Model} & \textbf{Model size} & \textbf{Pretraining tokens} & \textbf{ARC-c} & \textbf{HellaSwag} & \textbf{MMLU} & \textbf{TruthfulQA-mc2} & \textbf{WinoGrande} & \textbf{Average}\\
            \midrule[1.25pt]
            \gc{Cerebras-GPT \cite{dey2023cerebras}} & \gc{0.26 B} & \gc{5.1 B} & \gc{22.01} & \gc{28.99} & \gc{\bfseries 26.83} & \gc{\textbf{45.98}} & \gc{52.49} & \gc{35.26} \\
             \gc{OPT \cite{zhang2022opt}} & \gc{0.35 B} & \gc{0.2 T} & \gc{23.55} & \gc{36.73} & \gc{26.02} & \gc{40.83} & \gc{52.64} & \gc{35.95} \\
             \model (Ours) & 0.27 B & 1.5 T & \textbf{27.65} & \textbf{47.15} & 25.72 & 39.24 & \textbf{53.83} & \textbf{38.72} \\
             \midrule
             \gc{Pythia \cite{biderman2023pythia}} & \gc{0.41 B} & \gc{0.3 T} & \gc{24.83} & \gc{41.29} & \gc{25.99} & \gc{\bfseries 40.95} & \gc{54.38} & \gc{37.49} \\
             MobiLlama \cite{thawakar2024mobillama} & 0.50 B & 1.3 T  & 29.52 & 52.75 & \textbf{26.09} & 37.55 & 56.27 & 40.44\\
             \model (Ours) & 0.45 B & 1.5 T & \textbf{30.20} & \textbf{53.86} & 26.01 & 40.18 & \textbf{57.22} & \textbf{41.50} \\
             \midrule
             MobiLlama \cite{thawakar2024mobillama} & 0.80 B & 1.3 T & 30.63 & 54.17 & 25.2 & 38.41 & 56.35 & 40.95\\
             Pythia \cite{biderman2023pythia} & 1.40 B & 0.3 T & 32.68 & 54.96 & 25.56 & \textbf{38.66} & 57.30 & 41.83 \\
             MobiLlama \cite{thawakar2024mobillama} & 1.26 B & 1.3 T & 34.64 & 63.27 & 23.87 & 35.19 & 60.77 & 43.55\\
             OLMo \cite{groeneveld2024olmo} & 1.18 B & 3.0 T & 34.47 & 63.81 & 26.16 & 32.94 & 60.46 & 43.57\\
             \model (Ours) & 1.08 B & 1.5 T & \textbf{36.69} & \textbf{65.71} & \textbf{27.05} & 36.98 & \textbf{63.22} & \textbf{45.93}\\
             \midrule
             \model (Ours) & 3.04 B & 1.5 T & 42.24 & 73.28 & 26.76 & 34.98 & 67.25 & 48.90 \\
             \bottomrule[1.5pt]
        \end{tabular}
    }
    \caption{Results on OpenLLM Leaderboard tasks with respect to the metrics defined in \cref{tab:openllm}.}
    \label{tab:openllm_results}
\end{subtable}
\vfill
\begin{subtable}[c]{2\columnwidth}
    \centering
    \resizebox{\columnwidth}{!}{
\begin{tabular}{l|cc|cccccccc|c}
\toprule[1.5pt]
\textbf{Model} & \textbf{Model size} & \textbf{Pretraining tokens} & \textbf{ARC-c} & \textbf{CrowS-Pairs} & \textbf{HellaSwag} & \textbf{MMLU}  & \textbf{PIQA}  & \textbf{RACE}  & \textbf{TruthfulQA} & \textbf{WinoGrande} & \textbf{Average} \\ 
\midrule[1.25pt]
\model (Ours) & 0.27 B & 1.5 T & 27.65 & 66.79       & 47.15     & 25.72 & 69.75 & 30.91 & 39.24      & 53.83      & 45.13   \\
\midrule
MobiLlama \cite{thawakar2024mobillama}  & 0.50 B  & 1.3 T & 29.52 & 65.47       & 52.75     & \textbf{26.09} & 71.11 & 32.15 & 37.55      & 56.27      & 46.37   \\
\model (Ours) & 0.45 B & 1.5 T & \textbf{30.20}  & \textbf{68.63}       & \textbf{53.86}     & 26.01 & \textbf{72.31} & \textbf{33.11} & \textbf{40.18}  & \textbf{57.22}  & \textbf{47.69}   \\
\midrule
MobiLlama \cite{thawakar2024mobillama} & 0.80 B & 1.3 T  & 30.63 & 66.25       & 54.17     & 25.2  & 73.18 & 33.68 & \textbf{38.41} & 56.35      & 47.23   \\
MobiLlama \cite{thawakar2024mobillama} & 1.26 B & 1.3 T & 34.64 & 70.24       & 63.27     & 23.87 & 74.81 & 35.02 & 35.19      & 60.77      & 49.73   \\
OLMo \cite{groeneveld2024olmo} & 1.18 B & 3.0 T & 34.47 & 69.95       & 63.81     & 26.16 & 75.14 & \textbf{36.75} & 32.94      & 60.46      & 49.96   \\
\model (Ours) & 1.08 B & 1.5T & \textbf{36.69} & \textbf{71.74}  & \textbf{65.71}     & \textbf{27.05} & \textbf{75.57} & 36.46 & 36.98      & \textbf{63.22}      & \textbf{51.68}   \\
\midrule
\model (Ours) & 3.04 B & 1.5 T & 42.24 & 73.29       & 73.28     & 26.76 & 78.24 & 38.76 & 34.98      & 67.25      & 54.35   \\
\bottomrule[1.5pt]
\end{tabular}
}
\caption{Results on LLM360 tasks with respect to the metrics defined in \cref{tab:llm360}.}
\label{tab:llm360_res}
\end{subtable}
\caption{\textbf{Comparison of \model with publicly available LLMs across various evaluation frameworks.}. We chose MobiLlama and OLMo as our baselines because they are pre-trained on public datasets using a similar or larger number of tokens. We evaluate \model, MobiLlama, and OLMo using the same LM evaluation harness version. Results for other models in \cref{tab:res_std_metrics} and \cref{tab:openllm_results} are taken from their official GitHub repositories and the OpenLLM leaderboard \cite{open-llm-leaderboard}, respectively. Best task accuracy for each model category is highlighted in bold. Models pre-trained with less data are highlighted in \gc{gray} color.
}
\label{tab:results}
\end{table*}

\section{Experimental Results}
\label{sec:results}

\paragraph{Pre-training results.} We evaluate the performance of \model on zero-shot and few-shot settings (\cref{tab:eval_metrics}). We compare \model with publicly available LLMs, namely PyThia \cite{biderman2023pythia}, Cerebras-GPT \cite{dey2023cerebras}, TinyLlama \cite{zhang2024tinyllama}, OpenLM \cite{openlm}, MobiLlama \cite{thawakar2024mobillama}, and OLMo \cite{groeneveld2024olmo}. The works most closely related to ours are MobiLlama and OLMo. These models are trained on comparable dataset mixtures, with similar or larger number of pre-training tokens.

In \cref{fig:train_prog}, the accuracy of \model is plotted against training iterations for 7 standard zero-shot tasks. We observe an overall increase in accuracy with longer training durations across most tasks. Additionally, the checkpoint obtained by averaging the last five checkpoints, collected at intervals of 5000 iterations, demonstrates comparable or slightly better accuracy compared to the final checkpoint obtained after 350k iterations. This improvement is likely due to noise reduction through weight averaging. Consequently, we use the averaged checkpoint for our main evaluations in \cref{tab:results}, instruction tuning experiments in \cref{tab:instruct-results}, and parameter-efficient tuning experiments in \cref{tab:peft-results}.

The results in \cref{tab:results} span across various evaluation frameworks, and highlights \model's effectiveness over existing methods. For instance, an \model variant with 1.1 billion parameters achieves 1.28\% (\cref{tab:res_std_metrics}), 2.36\% (\cref{tab:openllm_results}), and 1.72\% (\cref{tab:llm360_res}) higher accuracy compared to OLMo with 1.2 billion parameters. Remarkably, \model achieves this level of accuracy while using $2\times$ less pre-training data.

\paragraph{Instruction tuning results.} We use the cleaned variant of UltraFeedback~\cite{cui2023ultrafeedback,notus2023} dataset that consists of 60k prompts for instruction tuning. We do instruction tuning using Alignment Handbook library~\cite{alignment_handbook2023}. For optimization, we use either the statistical rejection sampling method~\cite{liuStatisticalRejectionSampling2024} or the direct preference optimization method~\cite{rafailovDirectPreferenceOptimization2023}. These sampling method details along with other hyper-parameters and fine-tuning details are given in \cref{sec:app_inst_hyper}. 

\cref{tab:instruct-results} shows that instruction tuning consistently improves \model's average accuracy by 1-2\% across different evaluation frameworks.

\begin{table*}[t!]
\begin{subtable}[c]{2\columnwidth}
    \centering
    \resizebox{0.85\columnwidth}{!}{
\begin{tabular}{@{}cc|ccccccc|c}
  \toprule[1.5pt]
  \textbf{Model Size}     & \textbf{Instruction Tuned?} & \textbf{ARC-c} & \textbf{ARC-e} & \textbf{BoolQ} & \textbf{HellaSwag} & \textbf{PIQA}  & \textbf{SciQ}  & \textbf{WinoGrande} & \textbf{Average} \\ \midrule[1.25pt]
  \multirow{2}{*}{0.27 B}  & \xmark                      & 26.45          & 45.08          & \textbf{53.98} & 46.71              & 69.75          & \textbf{84.70} & \textbf{53.91}      & 54.37            \\
                          & \cmark                      & \textbf{30.55} & \textbf{46.68} & 48.56          & \textbf{52.07}     & \textbf{70.78} & 84.40          & 52.72               & \textbf{55.11}   \\ \midrule
  \multirow{2}{*}{0.45 B}  & \xmark                      & 27.56          & 48.06          & 55.78          & 53.97              & 72.31          & 87.20          & 58.01               & 57.56            \\
                          & \cmark                      & \textbf{30.38} & \textbf{50.00} & \textbf{60.37} & \textbf{59.34}     & \textbf{72.63} & \textbf{88.00} & \textbf{58.96}      & \textbf{59.95}   \\ \midrule
  \multirow{2}{*}{1.08 B} & \xmark                      & 32.34          & \textbf{55.43} & 63.58          & 64.81              & \textbf{75.57} & \textbf{90.60} & 61.72               & 63.44            \\
                          & \cmark                      & \textbf{37.97} & 52.23          & \textbf{70.00} & \textbf{71.20}     & 75.03          & 89.30          & \textbf{62.75}      & \textbf{65.50}   \\ \midrule
  \multirow{2}{*}{3.04 B} & \xmark                      & 35.58          & 59.89          & 67.40          & 72.44              & 78.24          & \textbf{92.70} & 65.51               & 67.39            \\
                          & \cmark                      & \textbf{39.42} & \textbf{61.74} & \textbf{68.17} & \textbf{76.36}     & \textbf{79.00} & 92.50          & \textbf{66.85}      & \textbf{69.15}   \\ \bottomrule[1.5pt]
\end{tabular}
}
\caption{Results on zero-shot tasks with respect to the metrics defined in \cref{tab:std_metrics}. }
\label{tab:res_std_metrics_ist}
\end{subtable}
\vfill
\begin{subtable}[c]{2\columnwidth}
    \centering
    \resizebox{0.75\columnwidth}{!}{
\begin{tabular}{@{}cc|ccccc|c}
  \toprule[1.5pt]
  \textbf{Model Size}     & \textbf{Instruction Tuned?} & \textbf{ARC-c} & \textbf{HellaSwag} & \textbf{MMLU}  & \textbf{TruthfulQA} & \textbf{WinoGrande} & \textbf{Average} \\ \midrule[1.25pt]
  \multirow{2}{*}{0.27 B}  & \xmark                      & 27.65          & 47.15              & 25.72          & \textbf{39.24}      & \textbf{53.83}      & 38.72            \\
                          & \cmark                      & \textbf{32.51} & \textbf{51.58}     & \textbf{26.70} & 38.72               & 53.20               & \textbf{40.54}   \\ \midrule
  \multirow{2}{*}{0.45 B}  & \xmark                      & 30.20          & 53.86              & \textbf{26.01} & 40.18               & 57.22               & 41.50            \\
                          & \cmark                      & \textbf{33.53} & \textbf{59.31}     & 25.41          & \textbf{40.48}      & \textbf{58.33}      & \textbf{43.41}   \\ \midrule
  \multirow{2}{*}{1.08 B} & \xmark                      & 36.69          & 65.71              & \textbf{27.05} & 36.98               & 63.22               & 45.93            \\
                          & \cmark                      & \textbf{41.55} & \textbf{71.83}     & 25.65          & \textbf{45.95}      & \textbf{64.72}      & \textbf{49.94}   \\ \midrule
  \multirow{2}{*}{3.04 B} & \xmark                      & 42.24          & 73.28              & \textbf{26.76} & 34.98               & 67.25               & 48.90            \\
                          & \cmark                      & \textbf{47.70} & \textbf{76.87}     & 24.80          & \textbf{38.76}      & \textbf{67.96}      & \textbf{51.22}   \\ \bottomrule[1.5pt]
\end{tabular}
    }
    \caption{Results on OpenLLM Leaderboard tasks with respect to the metrics defined in \cref{tab:openllm}.}
\end{subtable}
\vfill
\begin{subtable}[c]{2\columnwidth}
    \centering
    \resizebox{0.85\columnwidth}{!}{
\begin{tabular}{@{}cc|cccccccc|c}
  \toprule[1.5pt]
  \textbf{Model Size}     & \textbf{Instruction Tuned?} & \textbf{ARC-c} & \textbf{CrowS-Pairs} & \textbf{HellaSwag} & \textbf{MMLU}  & \textbf{PIQA}  & \textbf{RACE}  & \textbf{TruthfulQA} & \textbf{WinoGrande} & \textbf{Average} \\ \midrule[1.25pt]
  \multirow{2}{*}{0.27 B}  & \xmark                      & 27.65          & \textbf{66.79}       & 47.15              & 25.72          & 69.75          & 30.91          & \textbf{39.24}      & \textbf{53.83}      & 45.13            \\
                          & \cmark                      & \textbf{32.51} & 66.01                & \textbf{51.58}     & \textbf{26.70} & \textbf{70.78} & 33.78          & 38.72               & 53.20               & \textbf{46.66}   \\  \midrule
  \multirow{2}{*}{0.45 B}  & \xmark                      & 30.20          & \textbf{68.63}       & 53.86              & \textbf{26.01} & 72.31          & 33.11          & 40.18               & 57.22               & 47.69            \\
                          & \cmark                      & \textbf{33.53} & 67.44                & \textbf{59.31}     & 25.41          & \textbf{72.63} & \textbf{36.84} & \textbf{40.48}      & \textbf{58.33}      & \textbf{49.25}   \\  \midrule
  \multirow{2}{*}{1.08 B} & \xmark                      & 36.69          & \textbf{71.74}       & 65.71              & \textbf{27.05} & \textbf{75.57} & 36.46          & 36.98               & 63.22               & 51.68            \\
                          & \cmark                      & \textbf{41.55} & 71.02                & \textbf{71.83}     & 25.65          & 75.03          & \textbf{39.43} & \textbf{45.95}      & \textbf{64.72}      & \textbf{54.40}   \\  \midrule
  \multirow{2}{*}{3.04 B} & \xmark                      & 42.24          & \textbf{73.29}       & 73.28              & \textbf{26.76} & 78.24          & \textbf{38.76} & 34.98               & 67.25               & 54.35            \\
                          & \cmark                      & \textbf{47.70} & 72.33                & \textbf{76.87}     & 24.80          & \textbf{79.00} & 38.47          & \textbf{38.76}      & \textbf{67.96}      & \textbf{55.73}   \\  \bottomrule[1.5pt]
\end{tabular}
}
\caption{Results on LLM360 tasks with respect to the metrics defined in \cref{tab:llm360}.}
\end{subtable}
\caption{\textbf{Instruction tuning improves \model's accuracy across different model sizes.}}
\label{tab:instruct-results}
\end{table*}

\paragraph{Parameter-efficient fine-tuning (PEFT) results.} We use the CommonSense reasoning training and evaluation setup \cite{Hu2023LLMAdaptersAA}. This setup provides 170k training samples across 8 multiple-choice datasets for PEFT studies with different methods, including LoRA \cite{Hu2021LoRALA} and DoRA \cite{Liu2024DoRAWL}. We integrate \model with these methods, and finetune the resulting model for three epochs using 8 NVIDIA H100 GPUs. \cref{tab:peft-results} shows that PEFT methods can be applied to \model. LoRA and DoRA deliver similar accuracy on  average across the given CommonSense reasoning datasets. 

\begin{table*}[t!]
    \centering
\resizebox{1.6\columnwidth}{!}{
\begin{tabular}{@{}cc|cccccccc|c}
  \toprule[1.5pt]
  \textbf{Model Size}     & \textbf{PEFT} & \textbf{ARC-c} & \textbf{ARC-e} & \textbf{BoolQ} & \textbf{HellaSwag} & \textbf{PIQA}  & \textbf{SIQA}  & \textbf{WinoGrande} & \textbf{OBQA} & \textbf{Average} \\ \midrule[1.25pt]
  \multirow{2}{*}{0.27 B} & LoRA & 
24.57 & 26.60 & 62.14 & 24.84 & 50.05 & 42.02 & 49.88 & 28.00 & 38.51 \\
  & DoRA &
26.19 & 28.07 & 62.20 & 25.22 & 50.11 & 44.42 & 50.12 & 31.20 & 39.69 \\
\midrule
  \multirow{2}{*}{0.45 B} & LoRA &
  28.67 & 29.88 & 62.29 & 25.85 & 52.39 & 49.59 & 50.91 & 33.20 & 41.60 \\
  & DoRA & 
  28.33 & 30.39 & 62.26 & 25.12 & 52.29 & 49.28 & 50.83 & 32.00 & 41.31 \\
\midrule
  \multirow{2}{*}{1.08 B} & LoRA & 
  45.14 & 61.11 & 61.77 & 77.95 & 72.31 & 69.70 & 61.64 & 59.20 & 63.60 \\
  & DoRA & 
  44.11 & 61.49 & 61.68 & 78.92 & 71.38 & 69.04 & 64.01 & 58.80 & 63.68 \\ 
\midrule
  \multirow{2}{*}{3.04 B} & LoRA & 
  46.93 & 66.25 & 62.48 & 81.22 & 75.19 & 70.62 & 65.51 & 58.20 & 65.80 \\
  & DoRA & 
  46.50 & 66.46 & 62.35 & 80.84 & 75.73 & 70.83 & 63.77 & 58.20 & 65.59 \\ 
  \bottomrule[1.5pt]
\end{tabular}
}
\caption{\textbf{\model with PEFT.} 
Both LoRA and DoRA demonstrate comparable performance when \model is finetuned on CommonSense reasoning benchmark. It's important to note that these \emph{fine-tuning results}, obtained using the evaluation setup of LLM-Adapters \cite{Hu2023LLMAdaptersAA}, differ from the results in \cref{tab:results,tab:instruct-results}. This is because the results in \cref{tab:results,tab:instruct-results} are obtained under zero- and few-shot settings using LM Evaluation Harness. \gc{Note that we did not use social interactions QA (SIQA; \cite{sap2019socialiqa}) and OpenBookQA (OBQA; \cite{mihaylov2018can}) in \cref{tab:results,tab:instruct-results} because of evaluation issues with LLama tokenizer in LM Evaluation Harness (see \cite{Hsu2024LMEvalIssue}).}}
\label{tab:peft-results}
\end{table*}

\section{Benchmarking}
\label{sec:benchmarking}

\paragraph{Hardware.} We benchmark on modern, consumer-grade hardware with BFloat16 as the data type. Specifically, CUDA benchmarks were performed on a workstation with an Intel i9-13900KF CPU, equipped with 64 GB of DDR5-4000 DRAM, and an NVIDIA RTX 4090 GPU with 24 GB of VRAM, running Ubuntu 22.04. PyTorch v2.2.2 \cite{Paszke_PyTorch_An_Imperative_2019} was used, with the most recent versions of models and the associated libraries. HuggingFace Transformers v4.39.3 \cite{wolf-etal-2020-transformers} was used to benchmark HuggingFace models. We did not use Torch Inductor for model compilation.

To benchmark \model models on the Apple silicon, we used an Apple MacBook Pro with an M2 Max system-on-chip and 64GiB of RAM, running macOS 14.4.1. We ported the code and the weights of \model to Apple MLX v0.10.0 \cite{mlx2023}. To maximize the throughput, lazy evaluation was used in MLX with 8 tokens evaluated at a time.

\paragraph{Evaluation.} We provide two separate measurements for token throughput (measured in terms of tokens processed per second): (1) prompt processing (pre-fill), and (2) token generation. Additionally, we also report the total combined throughput. We benchmark all models sequentially, and execute one full ``dry run" generating 1024 tokens for the first model, since we found that this significantly increases the throughput of generation for subsequent models. Before measurement for each individual model, we warm up the model by executing a single forward pass to allow the frameworks to perform further auto-tuning, if any. In all experiments, we use key-value caching and generate 1024 tokens in addition to the prompt tokens in all tests. Static key-value cache was used whenever supported. The same prompt was used for all runs, resulting in prompt lengths of 35-36 tokens (depending on the tokenizer).

\begin{table}[t!]
    \centering
    \begin{subtable}[c]{\columnwidth}
    \centering
    \resizebox{0.9\columnwidth}{!}{
        \begin{tabular}{lcrrr}
            \toprule[1.5pt]
            \multicolumn{1}{c}{\multirow{2}{*}{\textbf{Model}}} & \multicolumn{1}{c}{\multirow{2}{*}{\textbf{Model size}}} & \multicolumn{3}{c}{\textbf{Throughput (Tokens per second)}} \\
            \cmidrule[1.25pt]{3-5}
            & & \multicolumn{1}{c}{\textbf{Prompt}} & \multicolumn{1}{c}{\textbf{Generation}} & \multicolumn{1}{c}{\textbf{Total}}\\
            \midrule[1.25pt]
             OPT \cite{zhang2022opt} & 0.35 B & 6524.17 & 214.11 & 220.21 \\
             \model (Ours) & 0.27 B & 6427.27 & 159.67 & 165.85 \\
             \midrule
             MobiLlama \cite{thawakar2024mobillama} & 0.50 B & 3423.25 & 136.35 & 146.86 \\
             \model (Ours) & 0.45 B & 5211.35 & 128.46 & 133.42 \\
             \midrule
             MobiLlama \cite{thawakar2024mobillama} & 0.80 B & 4151.75 & 126.01 & 130.08 \\
             Pythia \cite{biderman2023pythia} & 1.40 B & 4501.85 & 139.65 & 143.83  \\
             MobiLlama \cite{thawakar2024mobillama} & 1.26 B & 4938.29 & 142.96 & 147.67 \\
             OLMo \cite{groeneveld2024olmo} & 1.18 B & 7151.65 & 203.40 & 209.26 \\
             \model (Ours) & 1.08 B & 3681.73 & 92.15 & 95.72 \\
             \midrule
             \model (Ours) & 3.04 B & 2712.56 & 70.11 & 72.82  \\
             \bottomrule[1.5pt]
        \end{tabular}
    }
    \caption{Results on NVIDIA CUDA / Linux.}
    \label{tab:cuda_perf_results}
\end{subtable}
\vfill
\begin{subtable}[c]{\columnwidth}
    \centering
    \resizebox{0.75\columnwidth}{!}{
\begin{tabular}{lrrr}
\toprule[1.5pt]
\multicolumn{1}{c}{\multirow{2}{*}{\textbf{Model}}} & \multicolumn{3}{c}{\textbf{Throughput (Tokens per second)}} \\ 
\cmidrule[1.25pt]{2-4}
& \multicolumn{1}{c}{\textbf{Prompt}} & \multicolumn{1}{c}{\textbf{Generation}} & \multicolumn{1}{c}{\textbf{Total}}\\
\midrule[1.25pt]
\model-0.27B & 1151.41 & 212.40 & 218.45         \\
\model-0.27B-4bit & 803.99 & 256.35 & 262.70         \\
\midrule
\model-0.45B & 910.61 & 147.26 & 151.57         \\
\model-0.45B-4bit & 883.19 & 197.81 & 203.16         \\
\midrule
\model-1.08B & 508.56 & 78.72 & 81.04         \\
\model-1.08B-4bit & 554.17 & 117.90 & 121.14         \\
\midrule
\model-3.04B-bf16 & 234.96 & 33.96 & 34.97         \\
\model-3.04B-bf16-4bit & 211.32 & 60.33 & 61.83         \\
\bottomrule[1.5pt]

\end{tabular}
}
\caption{Results for the MLX port on Apple macOS.}
\label{tab:mlx_perf_results}
\end{subtable}
\caption{\textbf{Benchmark measurements of \model compared to other similar LLMs in its class.}. On CUDA, we evaluate \model, MobiLlama, and OLMo using the CoreNet version of \model and HuggingFace for the other two. On macOS, we only provide results for the MLX version of \model.
}
\label{tab:results_benchmark}
\end{table}

\begin{table}[t!]
    \centering
    \resizebox{\columnwidth}{!}{
        \begin{tabular}{lcrrr}
            \toprule[1.5pt]
            \multicolumn{1}{c}{\multirow{2}{*}{\textbf{Model}}} & \multicolumn{1}{c}{\textbf{Normalization layer}} & \multicolumn{3}{c}{\textbf{Throughput (Tokens per second)}} \\
            \cmidrule[1.25pt]{3-5}
            & \multicolumn{1}{c}{\bfseries (\# Invocations per token)} & \multicolumn{1}{c}{\textbf{Prompt}} & \multicolumn{1}{c}{\textbf{Generation}} & \multicolumn{1}{c}{\textbf{Total}}\\
            \midrule[1.25pt]
            \multirow{2}{*}{OLMo} & LayerNorm (33) & 7151.65 & 203.40 & 209.26 \\
            & RMSNorm-Naive (33) & 5360.56 & 171.41 & 176.92 \\
            \midrule
            \multirow{3}{*}{\model (Ours)} & LayerNorm (113) & 4697.50 & 130.34 & 135.38 \\
            & RMSNorm-Naive (113) & 3681.73 & 92.15 & 95.72 \\
            & RMSNorm-Apex (113) & 4280.66 & 113.42 & 117.81 \\
            \bottomrule[1.5pt]
    \end{tabular}
    }
    \caption{\textbf{Normalization layers are a bottleneck.} The throughput of both OLMo-1.18B and \model-1.08B significantly decreases with the naive implementation of RMSNorm in PyTorch compared to highly optimized LayerNorm \cite{ba2016layer}. Although Apex's \cite{nvidia_apex} RMSNorm implementation leads to notable throughput improvements compared to the naive implementation, a considerable performance gap persists in comparison to LayerNorm. This highlights the substantial optimization potential for future endeavors. The number of invocations per token for each normalization layer is indicated next to the layer name in brackets.}
    \label{tab:norm_bottleneck}
\end{table}

\paragraph{Results.} \cref{tab:cuda_perf_results,tab:mlx_perf_results} shows the benchmarking results on GPU and MacBook Pro respectively. Despite \model's higher accuracy for a similar parameter count, we observe that it is slower than OLMo. While the primary focus of this study is reproducibility rather than inference performance, we did comprehensive profiling to understand the bottlenecks. Our analysis reveals that a significant portion of \model's processing time can be attributed to our naive implementation of RMSNorm (\cref{tab:norm_bottleneck}). Specifically, naive RMSNorm implementation results in many individual kernel launches each of which processes a small input, rather than a launch of a single, fused kernel, as would be the case with \textit{e.g.} LayerNorm. By replacing the naive RMSNorm with Apex's RMSNorm  \cite{nvidia_apex}, we observe a notable increase in \model's throughput.  However, a substantial performance gap persists compared to the models that use optimized LayerNorm, in part because (1) \model has 113 RMSNorm layers as compared to 33 LayerNorm layers in OLMo and (2) Apex's RMSNorm is not optimized for small inputs. To further illustrate the performance degradation attributable to RMSNorm, we replaced the LayerNorm in OLMo with RMSNorm, and observed a significant drop in generation throughput. In future work, we plan to explore optimization strategies to further improve the inference efficiency of \model.

\section{Conclusion}
This work releases \model, a decoder-only transformer-based open language model. The \model uses a layer-wise scaling method for efficient parameter allocation within the transformer model, resulting in improved accuracy compared to existing models. Additionally, we have made the entire framework open, including training logs, multiple checkpoints, pre-training configurations, and MLX inference code. This extensive release aims to empower and strengthen the open research community, facilitating future research efforts.

\section*{Author Contributions}

The \model project was led by Sachin Mehta, with additional lead contributions from Mohammad Rastegari and Peter Zatloukal. \model would not have been possible without the help of our many teammates and collaborators. We list author contributions below:

\begin{itemize}[leftmargin=*]
    \item[] \textbf{Pre-training dataset collection and tooling:} Sachin Mehta and Mohammad Sekhavat
    \item[] \textbf{Architecture design:} Sachin Mehta
    \item[] \textbf{Model training:} Sachin Mehta and Mohammad Sekhavat
    \item[] \textbf{Evaluation suite and tooling:} Sachin Mehta, Qingqing Cao, Mohammad Sekhavat, Mahyar Najibi, Maxwell Horton, and Iman Mirzadeh.
    \item[] \textbf{Huggingface integration:} Qingqing Cao
    \item[] \textbf{Instruction tuning:} Qingqing Cao
    \item[] \textbf{Parameter-efficient finetuning:} Maxwell Horton
    \item[] \textbf{Performance analysis and MLX conversion:} Chenfan Sun, Dmitry Belenko, and Mahyar Najibi
    \item[] \textbf{Code review, bug fixes, and  maintenance:} Sachin Mehta, Maxwell Horton, Mohammad Shekhavat, and Yanzi Jin
\end{itemize}

\section*{Acknowledgements}
We extend our gratitude to the following people for discussions and assistance: Farzad Abdolhosseini, David Harrison, Mehrdad Farajtabar, Fartash Faghri, Oncel Tuzel,  Hadipour Ansari,  Raviteja Vemulapalli, Aseem Wadhwa, Kumari Nishu, Danny Tormoen, Minsik Cho, Jason Ramapuram, Rich Moe, Arsalan Farooq, Dom L'Eplattenier, Mayank Goel, Hassan Babaie, Chong Wang, Ruoming Pang, Tom Gunter, Antonie Lin, Irina Belousova, and Joris Pelemans.

\section*{Broader Impact}
The release of \model models aims to empower and enrich the open research community by providing access to state-of-the-art language models. Trained on publicly available datasets, these models are made available without any safety guarantees. Consequently, there exists the possibility of these models producing outputs that are inaccurate, harmful, biased, or objectionable in response to user prompts. Thus, it is imperative for users and developers to undertake thorough safety testing and implement appropriate filtering mechanisms tailored to their specific requirements.

%%%%%%%%% REFERENCES
{\small
\bibliographystyle{ieee_fullname}
\bibliography{main}
}

\clearpage
\appendix

\section{Pre-training hyper-parameters}
\label{sec:appendix_pretrain_hyper}
The pre-training hyper-parameters for different \model configurations are given in \cref{tab:app_lm_arch}.

\begin{table}[h!]
    \centering
    \resizebox{0.99\columnwidth}{!}{
    \begin{tabular}{lcccc}
        \toprule[1.5pt]
        & \textbf{270M} & \textbf{450M} & \textbf{1.1B} & \textbf{3B}  \\
        \midrule[1.25pt]
        Dimension $d_{model}$ & 1280 & 1536 & 2048 & 3072 \\
        Num. of layers $N$ & 16 & 20 & 28 & 36  \\
        Head dimension $d_h$ & 64 & 64 & 64 & 128 \\
        $\alpha_{min}, \alpha_{max}$ (\cref{eq:layer_wise_scale}) & \multicolumn{4}{c}{0.5, 1.0} \\
        $\beta_{min}, \beta_{max}$ (\cref{eq:layer_wise_scale}) & \multicolumn{4}{c}{0.5, 4.0} \\
        Normalization layer & \multicolumn{4}{c}{RMSNorm} \\
        Positional embeddings & \multicolumn{4}{c}{RoPE} \\
        Attention variant & \multicolumn{4}{c}{Grouped query attention} \\
        Activation & \multicolumn{4}{c}{SwiGLU} \\
        Context length & \multicolumn{4}{c}{2048} \\
        Batch size (tokens) & \multicolumn{4}{c}{approx. 4M} \\
        Weight tying \cite{press2016using} & \multicolumn{4}{c}{yes} \\
        \midrule
        Warm-up iterations & \multicolumn{4}{c}{5,000} \\
        Training steps & \multicolumn{4}{c}{350,000} \\
        Warm-up init. LR & \multicolumn{4}{c}{0.000001} \\
        Max. LR & 0.0053 & 0.0039 & 0024 & 0.0012\\
        Min. LR & \multicolumn{4}{c}{10\% of the max. LR} \\
        Loss function & \multicolumn{4}{c}{Cross-entropy} \\
        Optimizer & \multicolumn{4}{c}{AdamW ($\beta_1$=0.9, $\beta_2$=0.95, $\epsilon=1.e-8$)} \\
        Weight decay & \multicolumn{4}{c}{0.1} \\
        \midrule
        Activation checkpointing & \xmark & \cmark & \cmark & \cmark \\
        FSDP & \xmark & \xmark & \xmark & \cmark \\
        GPUs & 128 & 128 & 128 & 128 \\
        GPU Type & A100 & H100 & A100 & H100 \\
        GPU Memory & 80 GB & 80 GB & 80 GB & 80 GB\\
        Training time (in days) & 3 & 3 & 11 & 13\\
        \bottomrule[1.5pt]
    \end{tabular}
    }
    \caption{Pre-training details for different variants of \model.}
    \label{tab:app_lm_arch}
\end{table}

\section{Hyper-parameters for instruction tuning}
\label{sec:app_inst_hyper}

We conducted a grid search to determine optimal values for the learning rate and training epochs. For the learning rate, we explored values in the range of [2e-5, 3e-5, 5e-5, 8e-5, 1e-4], while for training epochs, we investigated the range of [3, 5, 8, 10].  The final recipe selected is the one that yielded the highest average accuracy across various tasks as presented in \cref{tab:std_metrics} and \cref{tab:llm360}. 

We finetune all the models with BFloat16 as a data type. We use activation checkpointing along with gradient accumulation with a step size of two. We use the AdamW optimizer with default beta values. We use the cosine learning rate scheduler with a warm-up ratio of 0.1, and we set the weight decay to 0 and loss temperature beta to 0.01. We set the maximum context length to 1024 and maximum prompt length to 512. Other hyper-parameters are included in \cref{tab:inst_details}.

\begin{table}[ht!]
    \centering
    \resizebox{0.99\columnwidth}{!}{
    \begin{tabular}{lcccc}
        \toprule[1.5pt]
        & \textbf{270M} & \textbf{450M} & \textbf{1.1B} & \textbf{3B}  \\
        \midrule[1.25pt]
        Batch size  & \multicolumn{4}{c}{8} \\
        Training epochs & 5 & 8 & 5 & 10 \\
        Learning rate & 2e-5 & 3e-5 & 5e-5 & 1e-4 \\
        Loss function & hinge & hinge & sigmoid & hinge \\
        \midrule
        DeepSpeed Zero3 \cite{rasley2020deepspeed} & \xmark & \cmark & \cmark & \cmark \\
        GPUs & \multicolumn{4}{c}{8} \\
        GPU Type & A100 & A100 & A100 & A100 \\
        GPU Memory & 40 GB & 40 GB & 40 GB & 80 GB\\
        Training time (in hours) & 2.5 & 4.3 & 6.6 & 14.2 \\
        \bottomrule[1.5pt]
    \end{tabular}
    }
    \caption{Instruction tuning details for different variants of \model.}
    \label{tab:inst_details}
\end{table}

\end{document}